\documentclass[1p]{elsarticle}

\usepackage{lineno,hyperref}
\usepackage{lipsum}
\usepackage{bm,upgreek,amsmath, amsfonts, color, amsthm}
\usepackage[linesnumbered, boxed]{algorithm2e}
\SetKwComment{Comment}{$\triangleright$\ }{}
\usepackage[inline]{enumitem}
\usepackage{ragged2e}
\usepackage[font=footnotesize]{subfig}
\DeclareMathOperator{\Tr}{Tr}
\newcommand{\quotes}[1]{``#1''}
\usepackage{tkz-graph} 
\usepackage{rotating}
\usepackage{multirow}
\usepackage{amssymb}
\usetikzlibrary{positioning,shapes,snakes,arrows,arrows.meta, quotes}

\modulolinenumbers[5]
\theoremstyle{definition}
\newtheorem{definition}{Definition}

\makeatletter
\def\ps@pprintTitle{%
 \let\@oddhead\@empty
 \let\@evenhead\@empty
 \def\@oddfoot{}%
 \let\@evenfoot\@oddfoot}
\makeatother

\begin{document}

\begin{frontmatter}

\title{Stable specification search in structural equation model with latent variables}

\author[UII,RUN]{Ridho~Rahmadi}\corref{mycorrespondingauthor}
\author[RUN]{Perry~Groot}
\author[RUN]{Tom~Heskes}
\address[UII]{Department of Informatics, Universitas Islam Indonesia.}
\address[RUN]{Institute for Computing and Information Sciences, Radboud University Nijmegen, the Netherlands.}
\cortext[mycorrespondingauthor]{Corresponding author, email: \href{mailto:r.rahmadi@cs.ru.nl}{r.rahmadi@cs.ru.nl}, visiting address: Faculty of Science, Toernooiveld 212, 6525EC Nijmegen, the Netherlands.}

\begin{abstract}
In our previous study, we introduced stable specification search for cross-sectional data (S3C). It is an exploratory causal method that combines stability selection concept and multi-objective optimization to search for stable and parsimonious causal structures across the entire range of model complexities. In this study, we extended S3C to S3C-Latent, to model causal relations between latent variables. We evaluated S3C-Latent on simulated data and compared the results to those of PC-MIMBuild, an extension of the PC algorithm, the state-of-the-art causal discovery method. The comparison showed that S3C-Latent achieved better performance. We also applied S3C-Latent to real-world data of children with attention deficit/hyperactivity disorder and data about measuring mental abilities among pupils. The results are consistent with those of previous studies.
\end{abstract}


\end{frontmatter}

\sloppy

\section{Introduction}
In many empirical sciences, it is of great interest to identify causal relationships among entities that are  not measured directly. Such entities are called latent variables or factors. A latent variable is typically used to represent a rather general concept that influences multiple measured or observed variables obtained from, for example, a questionnaire designed by experts in the field \citep{silva2006learning}. An example from psychometrics is the fatigue catastrophizing scale (FCS) \citep{jacobsen1999relation}, that is used to assess a patient's tendency of catastrophizing, a cognitive process characterized by a lack of confidence and an expectation of negative outcomes \citep{sullivan1990relation}, as a response of experiencing fatigue. It is a 10-item scale, and each item therein, e.g., \quotes{I find myself expecting the worst when I am fatigued}, is rated on 5-point scales (1 means \emph{never true} and 5 means \emph{all of the time true}).

Domains involving both observed and latent variables can be modeled using a structural equation model (SEM) with latent variables (also called a general SEM in \citep{bollen1989}), that can be divided into two main components: the measurement model and the structural model. The measurement model captures relationships between the latent variables and their corresponding observed variables, often called indicators. The structural model represents relationships among latent variables. A few approaches have been developed to discover the measurement model, such as factor analysis or a more advanced alternative introduced by \citep{silva2006learning}. 
Recently, a new method has been introduced to model causal relations between latent variables in the structural model \citep{ruifei2018copulafactor}.

To estimate underlying causal mechanisms is, however, a complex problem. First, given $p$ variables in the data, there will be $3^{p(p-1)/2}$ possible recursive structural models (described in Section~\ref{Section:SEMLatent})\citep{harwood2002genetic}.
Thus, even a modest number of variables implies an immense number of possible models. A typical SEM procedure starts with a hypothesized model, scores the model, and realizes a few model refinements (often called a specification search) to improve the score \citep{long1983covariance}, resulting in only a small number of model evaluations. 
Such a specification search is not intended to infer causal structures, but rather to detect and correct specification error between a proposed and the true model \citep{maccallum1986specification}.
Second, model estimation is known to be notoriously unstable. That is, a slight fluctuation in the data can lead to a considerable change in the final model. This makes that typical approaches in literature, that are based solely on a single estimation, cannot robustly estimate a causal model. 

In \citep{rahmadi2017causality}, we introduced stable specification search for cross-sectional data (S3C). S3C is an exploratory causal method that uses a (non-general) SEM to represent causal models with only observed variables. S3C represents a causal model using a SEM with observed variables. In the present study, we extend S3C to S3C-Latent to model causal relationships among latent variables. S3C and S3C-Latent are designed to resolve the aforementioned problems of an immense number of possible models and inherent instability in model estimation. We search for causal relations on the structural model and assume that
the measurement model is given and pure, i.e., each indicator is related to only one latent variable \citep{silva2006learning}. In addition, we also allow demographic variables such as gender and age to be included in the structural model. 

We evaluate the performance of S3C-Latent on different schemes of simulation and compare the results with those of PC-MIMBuild, an extension of the PC algorithm \citep{spirtes2000causation}. The results demonstrate that S3C-Latent performs better. We also apply S3C-Latent to real-world data from different domains, and the results are consistent with those of earlier studies.
This paper is organized as follows. Section~\ref{Section:Background} describes SEMs with latent variables and the S3C procedure. Section \ref{Section:S3C-Latent} introduces the proposed method, S3C-Latent. Section \ref{Section:ResultandDiscussion} reports and discusses the results on simulated and real-world data. Finally, Section \ref{Section:Conclusion} gives the conclusions of this study.

\section{Background}
\label{Section:Background}
\subsection{Structural equation model with latent variables}
\label{Section:SEMLatent}
\subsubsection{Representation}
We use a representation of a SEM with latent variables as described in
\citep{joreskog1977structural,bollen1989}. 
The SEM consists of the structural model that represents causal relationships among latent variables, and the measurement model that represents relationships from latent to observed variables. 
In the literature, the latent variable is often called a factor, and the observed variable is often called an indicator, a manifest, or a proxy. In this paper, we use those terms interchangeably. The structural model reads
\begin{equation} \label{Eq:StrucModel}
	\bm{\eta} = \bm{B\eta} + \bm{\Gamma\xi} + \bm{\zeta},
\end{equation}
where $\bm{\eta}$ is an $m\times1$ vector of latent endogenous (effect) variables, $\bm{\xi}$ is an $n\times1$ vector of latent exogenous (cause) variables, $\bm{\zeta}$ is an $m\times1$ vector of disturbances on $\bm\eta$, $\bm{B}$ is an $m\times m$ matrix of coefficients among $\bm\eta$, and $\bm{\Gamma}$ is an $m\times n$ matrix of coefficients among $\bm\xi$. In addition, $\bm{\Phi}$ and $\bm{\Psi}$ denote the covariance matrices of $\bm{\xi}$ and of $\bm{\zeta}$, respectively. We assume that $\mathbb{E}(\bm\eta)=\mathbb{E}(\bm\xi)=\mathbb{E}(\bm\zeta)=0$, $\bm{\xi}$ uncorrelated with $\bm{\zeta}$, and that $(\bm{I-B})$ is nonsingular. In this study, we assume a recursive structural model without reciprocal causal relations. Such a model can be represented using a directed acyclic graph (DAG) \citep{spirtes2010introduction}. A DAG is a graph with a set of nodes $V$ and a set of directed edges $E$, in which no directed cycles (reciprocal) relations.

The measurement model represents influences from $\bm{\eta}$ to its observed variables, an $r\times 1$ vector $\bm{x}$, and from $\bm{\zeta}$ to its observed variables, a $q\times 1$ vector $\bm{y}$. The measurement model reads
\begin{equation}
\begin{aligned} \label{Eq:MeasureModel}
	\bm{x}&=\bm{\Lambda}_x\bm{\xi}+\bm{\delta} \\
	\bm{y}&=\bm{\Lambda}_y\bm{\eta}+\bm{\epsilon},
\end{aligned}
\end{equation}
where the $r\times n$ matrix $\bm{\Lambda}_x$ and $q\times m$ matrix $\bm{\Lambda}_y$ contain the structure coefficients associating latent variables and indicators, and the $r\times 1$ vector $\bm{\delta}$ and $q\times 1$ vector $  \bm{\epsilon}$ contain errors on the indicators. In addition, an $r\times r$ matrix $\bm{\Theta}_\delta$ and a $q\times q$ matrix $\bm{\Theta}_\epsilon$ are the covariance matrices of $\bm{\delta}$ and of $\bm{\epsilon}$, respectively.

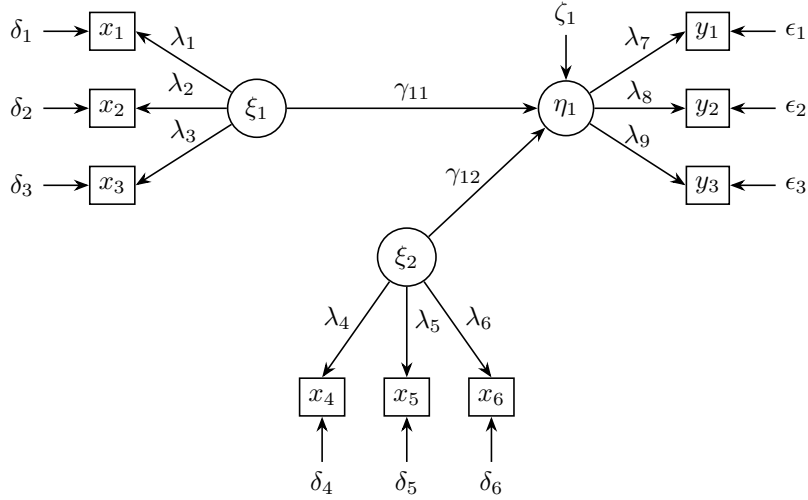
\begin{figure}
\center
\begin{tikzpicture}[
  transform shape, node distance=2cm,
  roundnode/.style={circle, draw=black, semithick, minimum size=7mm},
  squarednode/.style={rectangle, draw=black, semithick, minimum size=5mm},
  arrow/.style = {semithick,-Stealth},
  dotnode/.style={fill,inner sep=0pt,minimum size=2pt,circle} 
]

\node[roundnode] (latent1) {$\xi_1$};
\node[squarednode, left=1.2cm of latent1] (x2) {$x_2$}; 
\node[squarednode, above=0.5cm of x2]    (x1) {$x_1$};
\node[squarednode, below=0.5cm of x2]    (x3) {$x_3$};

\node[roundnode, right=3.3cm of latent1] (latent2) {$\eta_1$};
\node[squarednode, right=1.2cm of latent2] (y2) {$y_2$}; 
\node[squarednode, above=0.5cm of y2]    (y1) {$y_1$};
\node[squarednode, below=0.5cm of y2]    (y3) {$y_3$};

\node[roundnode, below right= of latent1] (latent3) {$\xi_2$};
\node[squarednode, below=1.2cm of latent3] (x5) {$x_5$}; 
 \node[squarednode, right=0.5cm of x5]    (x6) {$x_6$};
\node[squarednode, left=0.5cm of x5]    (x4) {$x_4$};

\foreach \i/\Yshift in {1/0,2/0,3/0}
{
	 \node[left=0.6cm of x\i] (d\i) {$\delta_\i$};
	 \draw[arrow] (d\i) -- (x\i);
    \draw[arrow] (latent1) -- node["$\lambda_\i$"{inner sep=1pt,yshift=\Yshift}]{} (x\i.east);
}

\foreach \i/\Xshift in {4/-7,5/8,6/8}
{
 \node[below=0.6cm of x\i] (d\i) {$\delta_\i$};
	 \draw[arrow] (d\i) -- (x\i);
    \draw[arrow] (latent3) -- node["$\lambda_\i$"{inner sep=-7pt,xshift=\Xshift}]{} (x\i.north);
}

\foreach \i in {1,2,3}
{
	 \node[right=0.6cm of y\i] (e\i) {$\epsilon_\i$};
	 \draw[arrow] (e\i) -- (y\i);
}

 \draw[arrow] (latent2) -- node["$\lambda_7$"{inner sep=1pt}]{} (y1.west);
 \draw[arrow] (latent2) -- node["$\lambda_8$"{inner sep=1pt, yshift=-3}]{} (y2.west);
 \draw[arrow] (latent2) -- node["$\lambda_9$"{inner sep=1pt, yshift=-2}]{} (y3.west);
 \draw[arrow] (latent1) -- node["$\gamma_{11}$"{inner sep=1pt, yshift=-2}]{} (latent2);
  \draw[arrow] (latent3) -- node["$\gamma_{12}$", pos=0.3]{} (latent2);

\node[above=0.6cm of latent2] (dist1) {$\zeta_1$};
\draw[arrow] (dist1) -- (latent2);

\end{tikzpicture}
\caption{An example of a SEM with three latent variables, each with three indicators.}
\label{SEMLatent}
\end{figure}

In general, a SEM procedure estimates a model-implied covariance matrix $\bm\Sigma(\bm\theta)$ and evaluates how closely it matches the sample covariance matrix $\mathbf{S}$. 
The $\bm\Sigma(\bm{\theta})$ is a function of model parameters $\bm{\theta}$ through
\begin{equation} \label{Equation:SigmaOfTheta} 
\begin{aligned} 
\bm\Sigma(\bm{\theta}) & = 
\begin{array} {lcl}
\begin{bmatrix}
	\bm\Sigma_{yy}(\bm{\theta}) & \bm\Sigma_{yx}(\bm{\theta}) \\[7pt] \bm\Sigma_{xy}(\bm{\theta}) & \bm\Sigma_{xx}(\bm{\theta})
\end{bmatrix}
\end{array},
 \\[7pt]
 \bm\Sigma_{yy}(\bm{\theta}) &= \bm{\Lambda}_y (\bm{I-B})^{-1}(\bm{\Gamma\Phi\Gamma}^\prime+\bm{\Psi})\begin{bmatrix} (\bm{I-B})^{-1}\end{bmatrix}^\prime\bm{\Lambda}_y^\prime+\bm{\Theta}_\epsilon,
\\[7pt]
\bm\Sigma_{xy}(\bm{\theta}) &= \bm{\Lambda}_x\bm{\Phi\Gamma}^\prime\begin{bmatrix}(\bm{I-B})^{-1}\end{bmatrix}^\prime\bm{\Lambda}_y^\prime,
\\[7pt]
\bm\Sigma_{xx}(\bm{\theta}) &= \bm{\Lambda}_x\bm{\Phi\Lambda}_x^\prime+\bm{\Theta}_\delta,
\end{aligned}
\end{equation}
where $\bm\Sigma_{yy}(\bm{\theta})$ is a covariance matrix of the indicators ${y}$ written as a function of the parameters $\bm{\theta}$. There are analogous definitions for $\bm\Sigma_{yx}(\bm{\theta})$, $\bm\Sigma_{xy}(\bm{\theta})$ and $\bm\Sigma_{xx}(\bm{\theta})$. The prime symbol indicates a matrix transpose.
Figure~\ref{SEMLatent} shows a SEM with three latent variables, each with three indicators.

\subsubsection*{Measurement model with ordinal indicators}
The SEM with latent variables represented by Equations~\ref{Eq:StrucModel} and \ref{Eq:MeasureModel} assumes normally distributed indicators $\bm{x}$ and $\bm{y}$. Based on \citep{olsson1982polyserial,drasgow1988polychoric}, 
this model can be extended to ordinal indicators $\hat{\bm{x}}$ and $\hat{\bm{y}}$ by discretizing, e.g., $\bm{x}$ into $w$ categories through
\begin{equation}
 \label{Eq:OrdinalModel}
	\hat{x}_i =\hat{x}_{ij} \qquad \text{if}\ \tau_{j-1} \leq x_i < \tau_j, \ j=1,\ldots, w, \ i=1,\ldots, r,
\end{equation}
where $\tau_j$ is \emph{threshold}, and for convenience, we define $\tau_0=-\infty$ and $\tau_w=+\infty$. The threshold $\tau_j$ and the values of $\hat{x}_{ij}$ are assumed to be strictly increasing, i.e., ${\tau_1<\tau_2<\ldots<\tau_{w-1}}$, and $\hat{x}_{i1}<\hat{x}_{i2}<\ldots <\hat{x}_{iw}$. There are analogous expressions for discretizing $\bm{y}$ into $w$ categories.

We assume that $\hat{\bm{x}}$ and $\hat{\bm{y}}$ are proxies of some continuous variables, and therefore use polychoric (between ordinal indicators) and polyserial (in the case of a mixture of continuous and ordinal indicators) correlations \citep{olsson1982polyserial,drasgow1988polychoric}, to form the matrix $\bm{S}$.

\subsubsection*{Markov equivalence class}
A structural model represented by a DAG has its corresponding Markov equivalence class or often called a completed partially DAG (CPDAG) \citep{chickering2002learning}. This implies that every probability distribution entailed by a structural model in a particular CPDAG, can also be derived by other structural models belonging to the same CPDAG.

\subsubsection{Identification and estimation}
\label{Section:IdenAndEst}
An integral part of our procedure will be the estimation of (the parameters of) SEMs. These parameters can only be estimated when the so-called identification conditions are satisfied. For the SEMs with latent variables that we consider in this paper, these conditions are as follows \citep{bollen1989,kenny1998data}.  
\begin{enumerate}[noitemsep]
\item \label{item:indicator} There are at least three or more indicators per latent variable.
\item Each row of $\bm{\Lambda}_x$ and $\bm{\Lambda}_y$ has only one nonzero element, i.e., an indicator cannot load on multiple latent variables (a pure model) \citep{silva2006learning}.
\item Each latent variable is scaled, e.g., by setting one factor loading $\lambda_{ij}$ of each latent $\zeta_j$ to $1$.
\item $\bm{\Theta_\delta}$ is diagonal. 
\end{enumerate}
Condition~\ref{item:indicator} can be relaxed to latent variables with less than three indicators as follows.
If there is a latent variable with two indicators, then the latent variable must have a causal relation with other latent variables \citep{bollen1989}.
If there is a latent variable with only one indicator, then the corresponding indicator error is set to zero \citep{kenny1998data}.

\subsubsection*{Estimation}

After a SEM satisfies the identification conditions, its model parameters $\bm\theta$ can be estimated through a maximum likelihood procedure, by minimizing a fitting function that reads
\begin{equation} \label{Equation:ML1}
\hat{\bm\theta} = \underset{\bm\theta}{\operatorname{argmin}}\, F_{ML}(\bm\theta),
\end{equation}
\begin{equation} \label{Equation:ML2}
 F_{ML}(\bm\theta)=\log|\bm{\Sigma}(\bm\theta)|+\Tr\{\bm{S\Sigma}^{-1}(\bm\theta)\}-\log|\bm{S}|-p.
\end{equation}
where $p$ is the number of observed variables, and $\bm{S}$ is the $p\times p$ sample covariance matrix of the observed variables.

\subsection{S3C procedure}
The S3C procedure can be divided into two main parts. The first part searches for the Pareto optimal causal models for the whole range of model complexities. This is realized by conducting a multi-objective optimization that mimics the idea of evolution \citep{deb2002fast}. The procedure is started with randomly generated SEMs that are scored based on two conflicting criteria: model fit and model complexity. 
The good models are characterized by those that fit the data well and have simple structures. They are selected to be parents to make new SEMs (children) in the next population of models by swapping and flipping model structures. This procedure is repeated many times, constituting many model refinements that lead to a better model population. At the end, the Pareto optimal models are obtained by filtering out those that are dominated by any other model in the population.

The second part of S3C aims to obtain SEM structures that are stable and simple (parsimonious). Adopting the idea of stability selection \citep{meinshausen2010stability}, S3C subsamples the data many times without replacement. For each of the subsets, S3C conducts the search that is described in the first part above, resulting in many Pareto optimal models. From those optimal models, S3C measures the edge and the causal path stability, that indicate the frequency of relations and the corresponding causal directions among pairs of variables across different model complexities (a precise definition of stability graphs is given in~\ref{Section:StabilityGraph}). Finally, with some thresholds S3C infers stable and parsimonious model structures (called \emph{relevant} structures) based on the edge and the causal path stability. For more details about the S3C procedure, we refer the readers to the original paper \citep{rahmadi2017causality}.

\section{Proposed method}
\label{Section:S3C-Latent}
Originally, S3C is designed to model causal relationships among observed variables. In the present study we extend S3C to S3C-Latent, to model causal relationships among latent variables. In particular, we use the representation of SEMs with latent variables as described in Section~\ref{Section:SEMLatent}.
As the basic idea of S3C is to search through many possible models, there is no restriction that forbids a cause $\xi_i$ in one model to be an effect $\eta_i$ in another model, and vice versa. The same condition implies to the corresponding indicators, that is, $x_i$ in one model can be $y_i$ in another model, and vice versa. However, there could be an exemption from this rule if one intends to incorporate \emph{prior knowledge}.

In practice, prior knowledge on the domain of interest may exist, e.g., results of previous studies that leads to a constraint on a particular causal relation. For example, based on an earlier study on patients with medically unexplained fatigue, it is known that the objective physical activity does not reduce the level of fatigue directly. In terms of a SEM with latent variables, this prior knowledge can be translated into a structural model with no directed edge from latent variable $O$ (denoting the objective physical activity) to latent variable $F$ (denoting the level of fatigue). Note that a directed path from $O$ to $F$ is still allowed, e.g., a path $O\to \dots\to F$ with any latent variables in between. S3C and S3C-Latent allow for incorporation of such prior knowledge.

In this study we further assume that the measurement model is given and pure. With respect to the data, we assume that samples have been generated independently and identically distributed (iid) from a linear Gaussian SEM.

Like S3C, to score a SEM, S3C-Latent uses the likelihood ratio test statistic \citep{joreskog1967some} (equal to $(N-1)F_{ML}$ from Equation (\ref{Equation:ML2}), with $N$ the sample size, which is often denoted $\chi_2$) to indicate the model fit, and the number of relations in the structural model (equals to the number of nonzero elements in matrices $\bm{B}$ and $\bm\Gamma$ in Equation~(\ref{Eq:StrucModel})), to indicate the model complexity. These scoring metrics are, however, not inherent to S3C-Latent, and can be replaced by different scoring metrics.

\subsection{S3C-Latent}
Let $D$ be the data set, $L=\{L_1,\ldots,L_n\}$ be a set of $n$ latent variables, $\Lambda$ be a matrix of factor loadings, and $C$ be prior knowledge. Figure~\ref{Algo:S3C-LatentPseudo} gives pseudocode of the S3C-Latent procedure.
\begin{figure}
\begin{algorithm}[H]
Ensure identification condititions $I$ fulfilled:\\
  \eIf{$\Lambda$ indicates that any latent variable $L_i\in L$ has indicators $<3$}{
   \eIf{the number of indicators $=2$}{ 
  		Set a relation between $L_i$ and one random latent $L_j\in L$ \\
  		Set one of the factor loadings on $L_i$ to $1$ 
   }{ 
   		Set the factor loading on $L_i$ to $1$ 
		\\
   		Set the error on the indicator to $0$
  }
   }{
   Set one of the factor loadings on each $L_i\in L$ to $1$ \\
  }
  Run S3C on $D$ with information of $L$ and satisfying $C$ and $I$
\end{algorithm}
 \caption{Algorithm of S3C-Latent}
\label{Algo:S3C-LatentPseudo}
\end{figure}

Lines $1$ to $11$ are to ensure that the measurement model identification conditions described in Section~\ref{Section:IdenAndEst} are fulfilled. In particular, Line $2$ checks if there are any latent variables $L_i \in L$ having less than $3$ indicators. If so, then for each $L_i $, Line $3$ is realized or Line $11$ otherwise. Line $3$ checks whether the number of indicators of $L_i$ is $2$ or $1$. In the case of $2$ indicators, S3C-Latent sets a relation between the latent variable $L_i$ and a random latent variable $L_j \in L$ ($L_i$ can be either a cause or an effect), and fixes one of its factor loadings to $1$ (Lines $4$ and $5$). In the case of $1$ indicator, S3C-Latent sets the factor loading of $L_i$ to $1$ and the indicator error to $0$ (Lines $7$ and $8$). Line $11$ is realized when all latent variables have at least $3$ indicators. In this case, one of the factor loadings on each latent variable is set to $1$.
Finally, Line $13$ runs S3C on data set $D$ with information of latent variables from $L$, satisfying any constraints in $C$, and fulfilling model identification conditions in $I$. By satisfying constraints in $C$ (if any), S3C-Latent ensures that all SEMs that are generated and refined, are consistent with the prior knowledge stated in $C$.

We implemented S3C-Latent as an \textsf{R} package and made it available online.\footnote{\url{https://github.com/rahmarid/S3C-Latent}} All of the source codes are provided, allowing interested users to modify the functionality.

\section{Result and discussion}
\label{Section:ResultandDiscussion}
\subsection{Application to simulated data}
To simulate different practical cases, we conducted different schemes of simulation by generating data from models with different number of latent variables, sample size, number of indicators, and types of indicators, e.g., continuous and ordinal. All of the \textsf{R} scripts used and the data sets generated for this simulation are available at the same repository of the \textsf{R} package.

\subsubsection{Model and data generation}
First, we randomly generated structural models of $n$ latent variables, by generating DAGs of $n$ nodes with a probability $s=2/(n-1)$ when relating a node (latent variable) to another node. Each DAG can be represented as an $n\times n$ adjacency matrix $\bm{A}$.
As the matrix $\bm{A}$ captures causal relations among latent variables $L=\{L_1,\ldots,L_n\}$, the matrices $\bm{B}$ and $\bm\Gamma$ in Equation (\ref{Eq:StrucModel}) are submatrices of $\bm{A}$. For each structural model, we then randomly formed two pure measurement models, where one has $3$ to $5$ indicators and the other one has $1$ to $4$ indicators per latent variable. This is realized by generating an $h \times n$ random matrix of factor loadings $\bm\Lambda$ for each structural model, where $h$ is the total number of indicators from $n$ latent variables. The matrices $\bm\Lambda_{x}$ and $\bm\Lambda_{y}$ in Equation (\ref{Eq:MeasureModel}) are submatrices of $\bm\Lambda$. With this procedure, we generated $20$ SEMs with $4$ and with $6$ latent variables.

Second, we simulated data sets of sizes $400$, $1000$, and $2000$ from each SEM with normal distribution, and then created the corresponding ordinal data sets by discretizing the values of the indicators into $2$ to $7$ ordered categories randomly. Taken together, we simulated $480$ data sets on which the performance of S3C-Latent has been evaluated. Table~\ref{tableSimScheme} summarizes the schemes of the simulation. Scheme C$_{3-5}$, for example, applied S3C-Latent to the data sets of continuous variables generated from SEMs with $3$ to $5$ indicators per latent variable.
Moreover,  we also applied PC-MIMBuild \citep{silva2006learning} on the same data sets, and then compared the results. PC-MIMBuild originates from the PC algorithm \citep{spirtes2000causation}, a state-of-the-art causal discovery method.

\begin{table}[!h]
\centering
\small
\caption{Types of indicators and the number of indicators per latent variable in different schemes of simulation.}
\begin{tabular}{|c|c|c|}
\hline
 Scheme & Type of indicators & Number of indicators \\ 
 \hline
 C$_{3-5}$ & Continuous & $3$ to $5$  \\
 C$_{1-4}$ & Continuous & $1$ to $4$  \\
 O$_{3-5}$ & Ordinal & $3$ to $5$  \\
 O$_{1-4}$ & Ordinal & $1$ to $4$  \\
 \hline
\end{tabular}
\label{tableSimScheme}
\end{table}

\subsubsection{Parameter settings}
A parameter setting of S3C-Latent consists of the number of subsets to draw (S), the number of iterations (I), the number of models to evaluate (P), crossover probability (C), and mutation probability (M). 
For applications to data sets generated from SEMs with $4$ latent variables, we set S $=25$, I $=30$, P $=50$, C $=0.45$, and M $=0.01$. For applications to data sets generated from SEMs with $6$ latent variables, we used the same parameter setting, except that we set I $=50$ and P $=100$.

For a fair comparison, we also drew $25$ subsets in the applications of PC-MIMBuild (similar to that in \citep{ramsey2010bootstrapping}) 
and measured the edge and the causal path stability as in \mbox{S3C-Latent}. We set the significance level for PC-MIMBuild in testing conditional independence to $0.01$.

\subsubsection{Evaluation performance}
We used the receiver operating characteristic (ROC) curve \citep{fawcett2004roc} to evaluate the performance of S3C-Latent and of PC-MIMBuild. Here, the true positive rate (TPR) and the false positive rate (FPR) are computed based on the CPDAG of the true model \citep{chickering2002learning}, while increasing the thresholds of stability. We measured the ROC for both the edge and the causal path stability. For example, a true positive means that a causal path with any length predicted by S3C-Latent or PC-MIMBuild actually exists in the CPDAG of the true model, while a false positive means that the predicted causal path does not exist in the CPDAG of the true model. 

Each simulation scheme comprises $20$ data sets of sizes $400$, $1000$, and $2000$, making it in total $60$ data sets. We measured the area under the curve (AUC) of each ROC and then computed the mean of the AUCs that S3C-Latent obtained over $20$ data sets (of the same sample size) and compared to that of PC-MIMBuild.

\subsubsection{Discussion}
\begin{figure}[!tb]
\centering
\includegraphics[width=1\textwidth]{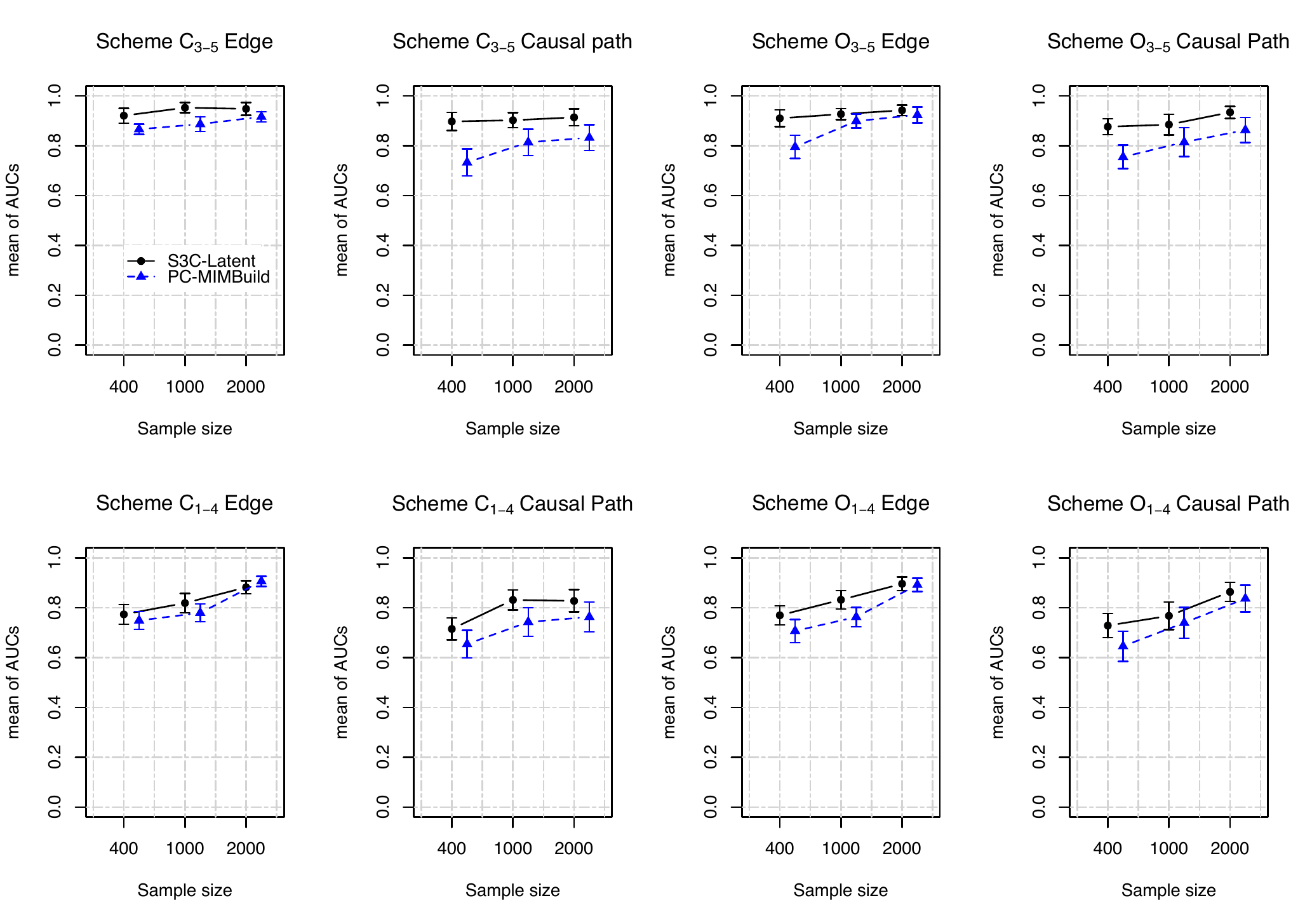} 
\caption{The mean AUC obtained by S3C-Latent (black-solid lines) and PC-MIMBuild (blue-dashed lines) over $20$ data sets generated from SEMs with $4$ latent variables, as a function of sample size. \quotes{Edge} indicates AUCs measured from the edge stability and \quotes{Causal Path} indicates those measured from the causal path stability. The plots in the top panels are results of simulations on the data sets generated from SEMs with $3$ to $5$ continuous (Scheme C$_{3-5}$) or ordinal (Scheme O$_{3-5}$) indicators per latent variable. The plots in the bottom panels are results of simulations on the data sets generated from SEMs with $1$ to $4$ continuous (Scheme C$_{1-4}$) or ordinal (Scheme O$_{1-4}$) indicators per latent variable.}
\label{Figure:Simfactor4}
\end{figure}

\begin{figure}[!tb]
\centering
\includegraphics[width=1\textwidth]{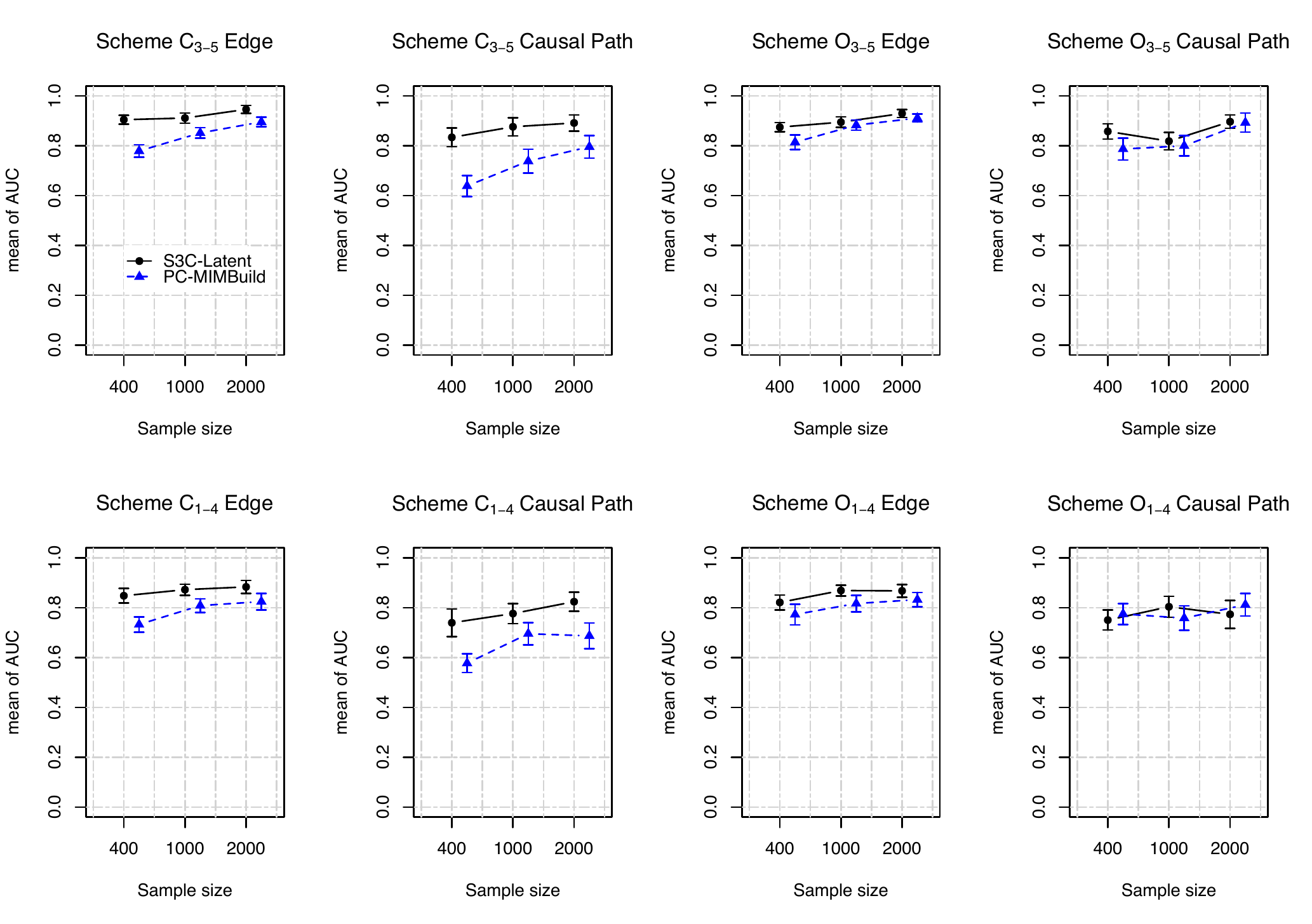} 
\caption{The mean AUC obtained by S3C-Latent (black-solid lines) and PC-MIMBuild (blue-dashed lines) over $20$ data sets generated from SEMs with $6$ latent variables, as a function of sample size. \quotes{Edge} indicates AUCs measured from the edge stability and \quotes{Causal Path} indicates those measured from the causal path stability. The plots in the top panels are results of simulations on the data sets generated from SEMs with $3$ to $5$ continuous (Scheme C$_{3-5}$) or ordinal (Scheme O$_{3-5}$) indicators per latent variable. The plots in the bottom panels are results of simulations on the data sets generated from SEMs with $1$ to $4$ continuous (Scheme C$_{1-4}$) or ordinal (Scheme O$_{1-4}$) indicators per latent variable.}
\label{Figure:Simfactor6}
\end{figure}

Figures~\ref{Figure:Simfactor4} and \ref{Figure:Simfactor6} show the plots of the mean AUC and the corresponding standard errors obtained by S3C-latent (black-solid lines) and PC-MIMBuild (blue-dashed lines), as a function of sample size. 
The top panels show results from Schemes C$_{3-5}$ and O$_{3-5}$, and the bottom panels show those from Schemes C$_{1-4}$ and O$_{1-4}$.

In all simulation schemes with $4$ latent variables, S3C-Latent achieved higher AUCs than those achieved by PC-MIMBuild. 
In a similar trend,  S3C-Latent obtained higher AUCs than those achieved by PC-MIMBuild on most simulation schemes with $6$ latent variables. The discrepancy of the results of S3C-Latent and of PC-MIMBuild becomes clearer in the case of continuous indicators.
Based on the results on simulations with $4$ and $6$ latent variables, the performance of S3C-Latent seems to be more robust compared to that of PC-MIMBuild across different sample sizes. 
In particular for small sample sizes, typical to many real-world applications, S3C Latent appears to be more robust than PC-MIMBuild.

Figure~\ref{Figure:SchemeABS3C} in \ref{Section:SchemeComp} shows comparisons of S3C-Latent's results in different schemes. As expected, in general the performance of S3C-Latent on data generated from SEMs with $3$ to $5$ indicators per latent variable is better than that of on data generated from SEMs with $1$ to $4$ indicators per latent variable.

\subsection{Application to real-world data}
\subsubsection{Attention deficit/hyperactivity disorder data}
For the first application, we applied S3C-Latent to a data set about attention deficit/hyperactivity disorder (ADHD) that was collected in \citep{van2012co}. 
The data set comprises observations on the children, among which $236$ have ADHD and $406$ belong to the control group. From all of the samples, $269$ are girls. There are four observed variables: gender, age, verbal intelligence quotient (VIQ), and performance intelligence quotient (PIQ). Moreover, there are $18$ questions distributed into three groups to indicate three latent variables, namely inattention (questions $1$-$9$), hyperactivity (questions $10$-$14$), and impulsivity (questions $15$-$18$).
We intend to model the causal relations between those four observed and the three latent variables. The missing data, that are assumed to be missing at random (MAR), are about $0.78\%$ and were imputed using expectation maximization (EM) \citep{honaker2011amelia}. We combined the imputation with our subsampling, resulting in different imputations across data subsets.
The parameter settings of S3C-Latent are S $=100$, I $=100$, P $=200$, C $=0.45$, and M $=0.01$. We included a prior knowledge that nothing causes gender.

\subsubsection{Discussion on ADHD result}
\label{Section:discussionADHD}
\begin{figure}[!htb]
\centering
\mbox{
\subfloat[]
{
\includegraphics[width=0.7\textwidth]{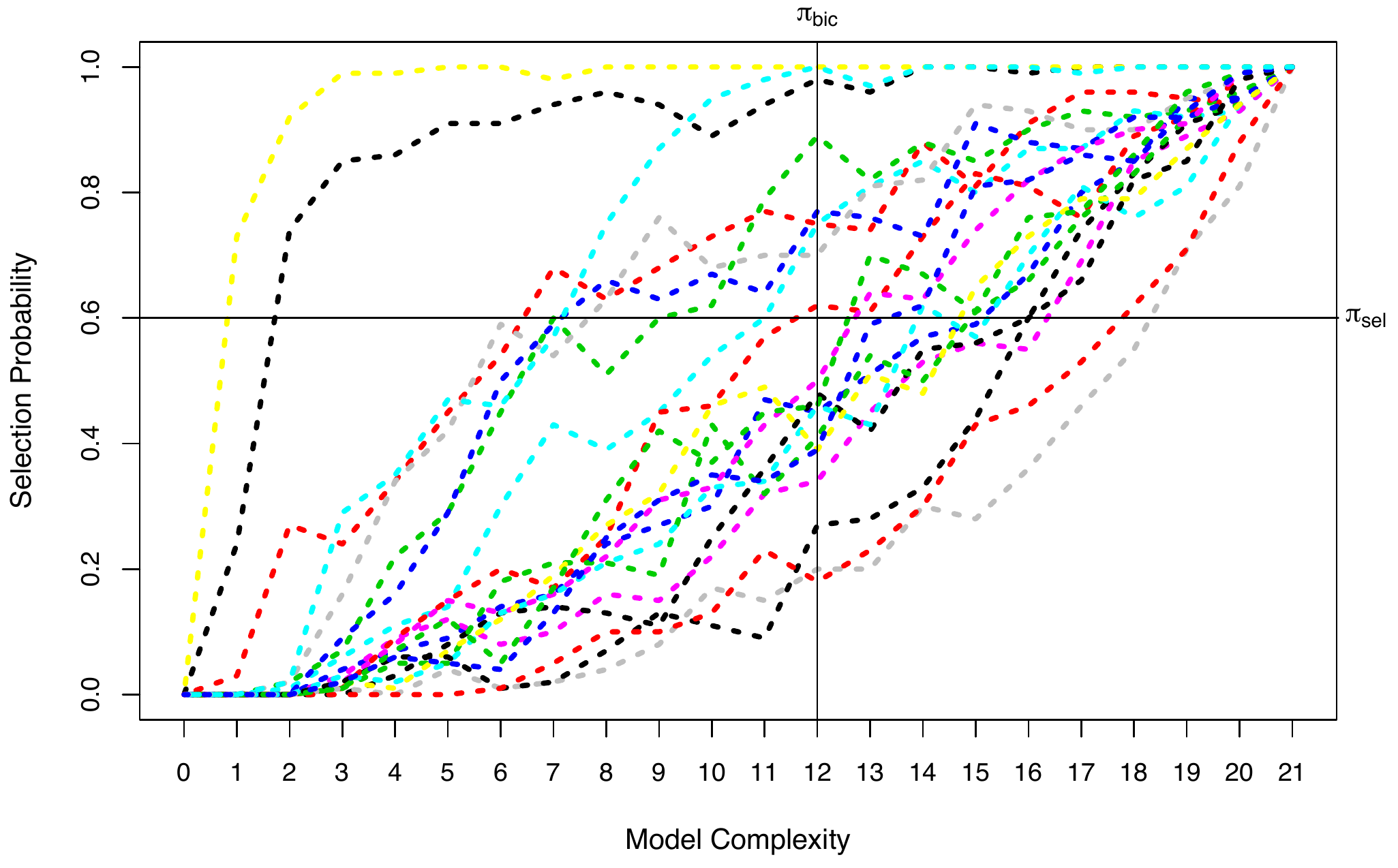}
\label{edgeStabADHD}
}}
\mbox{
\subfloat[]
{
\includegraphics[width=0.7\textwidth]{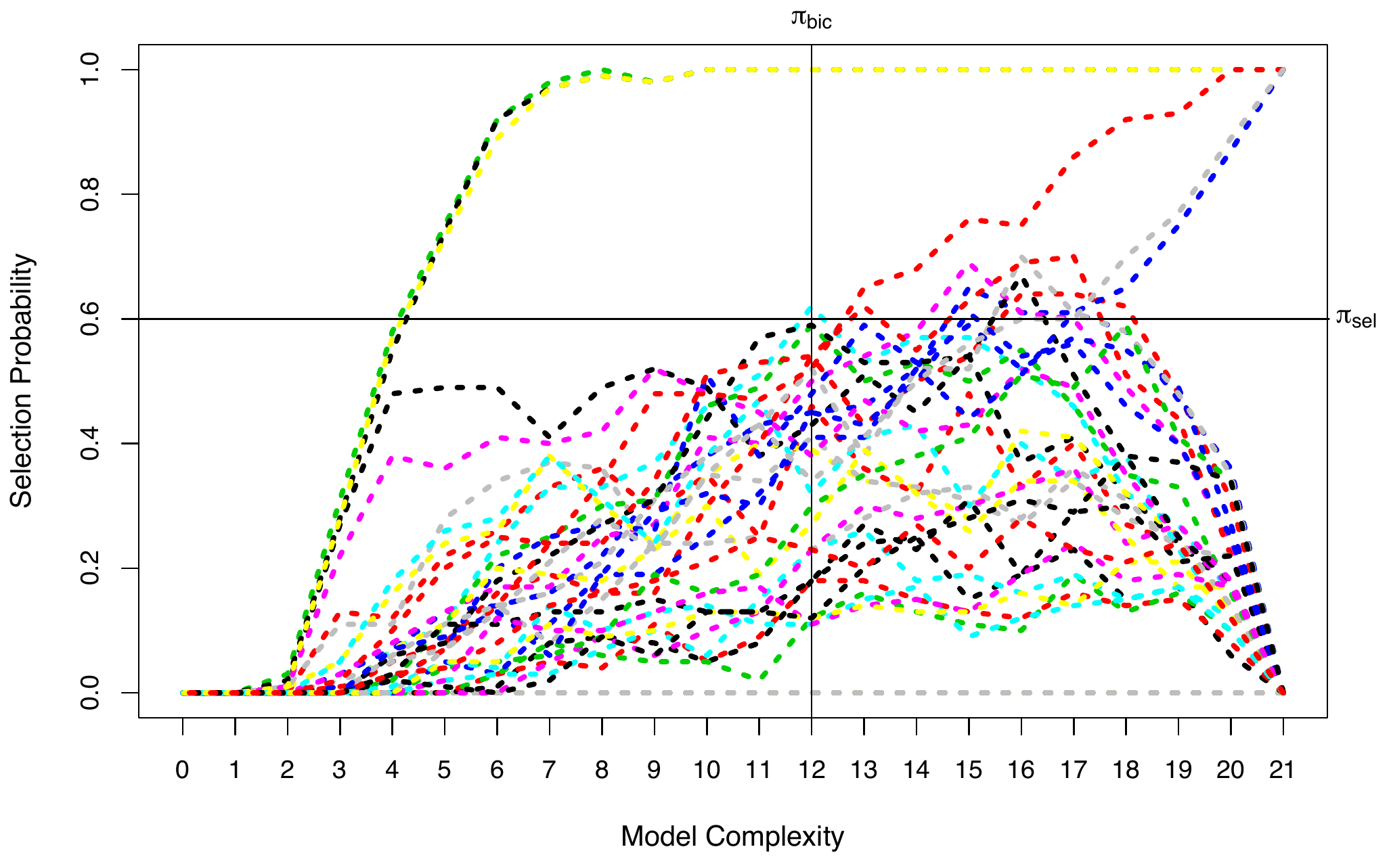}
\label{causalStabADHD} 
}}
\caption{The (a) edge and the (b) causal path stability graphs from the ADHD application. Each line indicates the frequency of relations between a pair of variables across different model complexities. The frequency is computed relative to the number of SEMs in each model complexity. The causal path stability only takes into account causal relations with any length, while the edge stability takes into account all relations regardless of the direction.}
\label{Figure:stabADHD}
\end{figure}

\begin{figure}[!htb]
\center
\footnotesize
\begin{tikzpicture}[
  transform shape, node distance=2cm,
  roundnode/.style={circle, draw=black, semithick, minimum size=14mm},
  squarednode/.style={rectangle, draw=black, semithick, minimum width=10mm, minimum height=7mm},
  squarednode2/.style={rectangle, draw=black, semithick, minimum width=15mm, minimum height=7mm},
  ellipsenode/.style={ellipse, draw=black, semithick, minimum size=5mm},
  arrow/.style = {semithick,-Stealth},
  dotnode/.style={fill,inner sep=0pt,minimum size=2pt,circle} 
]

\node[roundnode] (inatt) {Inatt};
\node[right=1.2cm of inatt] (qIn) {\vdots}; 
\node[squarednode, above=0.2cm of qIn]    (q1) {Q$_1$};
\node[squarednode, below=0.2cm of qIn]    (q9) {Q$_9$};
\draw[arrow, double distance=1.5pt] (inatt) -- node{} (q1.west);
\draw[arrow, double distance=1.5pt] (inatt) -- node{} (q9.west);

\node[roundnode, below=1.5cm of inatt] (hyper) {Hyper};
\node[right=1.2cm of hyper] (qHy) {\vdots}; 
\node[squarednode, above=0.2cm of qHy]    (q10) {Q$_{10}$};
\node[squarednode, below=0.2cm of qHy]    (q14) {Q$_{14}$};
\draw[arrow,double distance=1.5pt] (hyper) -- node{} (q10.west);
\draw[arrow,double distance=1.5pt] (hyper) -- node{} (q14.west);

\node[roundnode, below= 1.5cm of hyper] (impul) {Impul};
\node[right=1.2cm of impul] (qImp) {\vdots}; 
 \node[squarednode, below=0.2cm of qImp]    (q18) {Q$_{18}$};
\node[squarednode, above=0.2cm of qImp]    (q15) {Q$_{15}$};
\draw[arrow,double distance=1.5pt] (impul) -- node{} (q15.west);
\draw[arrow,double distance=1.5pt] (impul) -- node{} (q18.west);

\node[squarednode2, left=2.3cm of inatt] (gender) {Gender};
\node[squarednode2, left=1.7cm of gender] (age) {Age};
\node[squarednode2, left=2.3cm of hyper] (viq) {VIQ};
\node[squarednode2, left=1.7cm of viq] (piq) {PIQ};

\draw[arrow] (gender.east) -- node[above=-0.05, font=\scriptsize]{0.89/0.02} (inatt.west);
\draw[arrow] (gender.east) -- node[right,pos=0.17,font=\scriptsize]{0.76/0.04} (hyper.west);
\draw[arrow] (gender.east) -- node[left, pos=0.9,font=\scriptsize]{0.77/0.02} (impul.west);

\draw (inatt) edge[dashed] node[right,font=\scriptsize]{0.98}(hyper);
\draw (hyper) edge[dashed] node[right,font=\scriptsize]{1}(impul);
\draw (viq) edge[dashed] node[below, pos=0.22,font=\scriptsize]{0.62}(hyper);
\draw (viq) edge[dashed] node[above,font=\scriptsize]{1}(piq);
\draw (viq) edge[dashed] node[above=0.2,pos=0.15, font=\scriptsize]{0.75}(inatt);

\draw (inatt) edge[dashed, bend right=40] node[right,pos=0.7,font=\scriptsize]{0.77}(impul);

\end{tikzpicture}
\caption{A dashed undirected edge exhibits strong association where the causal direction cannot be determined from the data. A directed edge (arrow) with a single line indicates a causal relation and that with a double line indicates a factor loading. All edges, except the factor loadings, are annotated with a reliability score, i.e., the highest selection probability an edge has across the relevant region of the edge stability graph. In addition, all arrows are annotated with an estimate of the total causal effect (see~\ref{Section:CausalEffect} for more detail). For example, $0.89\slash 0.02$ indicates a reliability score of $0.89$ and a total causal effect of $0.02$.}
\label{Figure:modelADHD}
\end{figure}
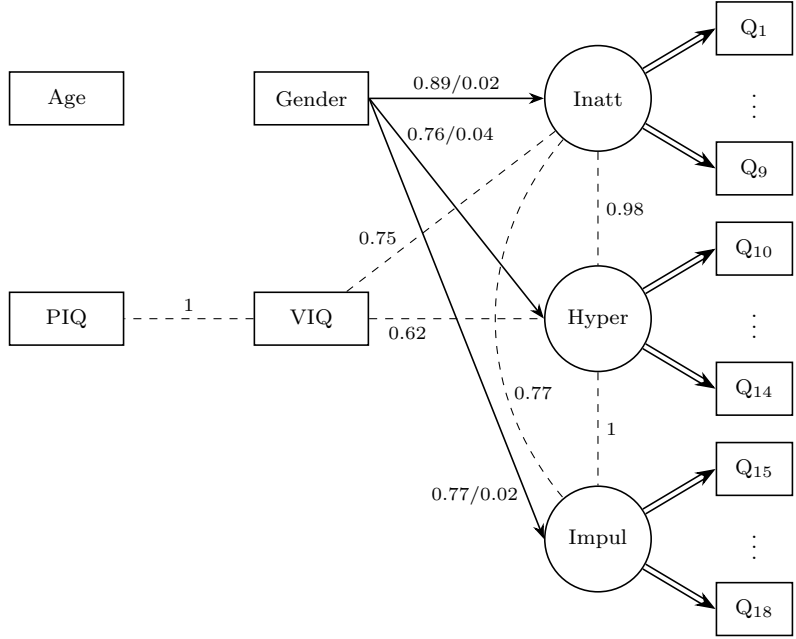

Figure~\ref{Figure:stabADHD} depicts the edge and the causal path stability graphs, with the $x$-axis indicating the model complexity and the $y$-axis indicating the selection probability. The threshold $\pi_{sel}$ (horizontal line) is set to $0.6$ and indicates that any model structure with selection probability equal to or greater than $\pi_{sel}$ is considered stable \citep{meinshausen2010stability}. The threshold $\pi_{bic}$ (vertical line) is set to the model complexity at which the minimum median of the BIC scores is found \citep{rahmadi2017causality}, which in this case was at $12$, indicating that any model structure with model complexity equal to or lesser than $\pi_{bic}$ is considered simple or parsimonious. Thus, the relevant model structures (stable and parsimonious) are those that pass through the top-left region of the stability graphs. The relevant model structures are then visualized with the steps described in \citep{rahmadi2017causality}, 
resulting in the graph shown in Figure~\ref{Figure:modelADHD}. A more detailed description on how to interpret the graph is given in the figure caption.

Based on Figure~\ref{Figure:modelADHD}, gender is found to influence inattention, hyperactivity, and impulsivity. This is in accordance with the studies of \citep{gaub1997gender,gershon2002meta,bauermeister2007adhd}
which exhibited that boys with ADHD tend to have higher symptom levels. Meta-analyses in population-based \citep{willcutt2000etiology} and clinically referred samples \citep{novik2006influence} suggested that boys are more likely to meet the Diagnostic and Statistical Manual of Mental Disorders (DSM) criteria for ADHD than girls.

VIQ is found associated with inattention and with hyperactivity, and that matches the studies of \citep{andreou2005verbal,frazier2004meta} 
which reported that children with ADHD have significantly lower VIQ compared to children without ADHD. The relation between VIQ and PIQ is expected as both constitute an assessment of the subject's intelligence. 

The association between inattention and hyperactivity is likely due to the direct causal relations from gender to both variables. A similar relation was found in the studies of \citep{sokolova2017causal, sokolova2016statistical} 
which exhibited that inattention influences hyperactivity. S3C-Latent found that the causal path stability from inattention to hyperactivity is about $0.4$. Having more samples might increase the stability (e.g., $>0.6$), so as to justify the same causal direction.

The relation between hyperactivity and impulsivity (as also found by \citep{sokolova2017causal}) 
is also likely due to the direct causal relations from gender to both variables. In fact, in most studies hyperactivity and impulsivity are regarded as one combined feature.

The association between inattention and impulsivity possibly follows from the direct relations between inattention and impulsivity with hyperactivity. This association was also indicated by \citep{sokolova2017causal}.
The associations between inattention, hyperactivity, and impulsivity are sensible as the three constitute the ADHD symptom. In addition, S3L did not find any relation between age with other variables.

\subsubsection{Holzinger data}
For the second application, we considered a well-known data set in the psychometric domain \citep{holzinger1939study},
obtained from an \textsf{R} package called \texttt{MBESS} $4.4.3$ \citep{kelley2007methods}.
The aim of the study was to learn factor pattern for mental abilities among two groups of pupils with different biological and cultural backgrounds \citep{douglas1940Factor}. For this purpose, $301$ students ($155$ females) from grades $7$ and $8$ were given $24$ tests, measuring spatial, verbal, mental speed, memory, and mathematical ability.  More details about the tests can be obtained from the package documentation. Our interest is to model causal relations among those latent variables including gender. We include a prior knowledge that nothing causes gender. We set the S3C-Latent parameters S $=100$, I $=80$, P $=150$, C $=0.45$, and M $=0.01$.

\subsubsection{Discussion on Holzinger result}
\begin{figure}[!htb]
\centering
\mbox{
\subfloat[]
{
\includegraphics[width=0.7\textwidth]{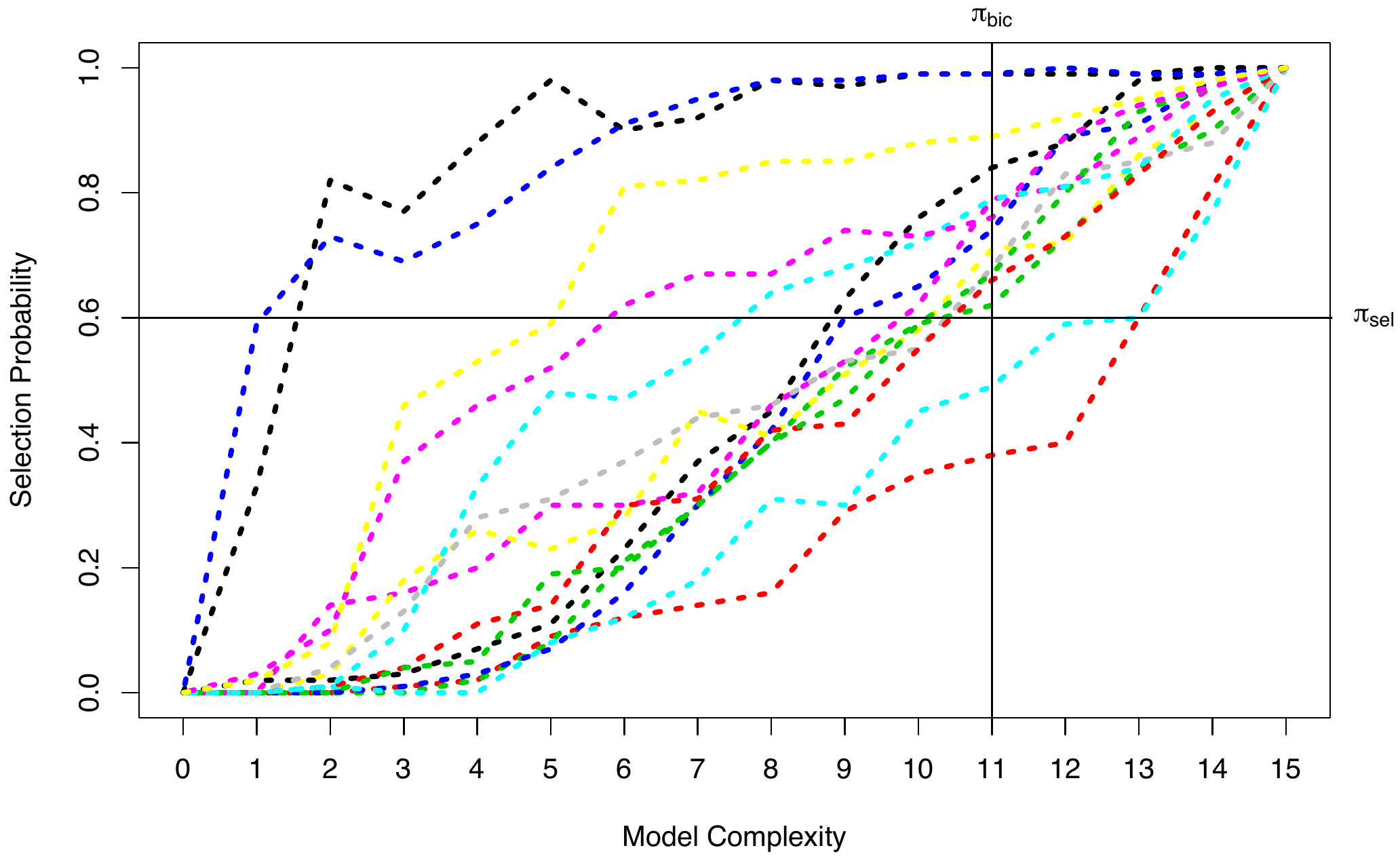}
\label{edgeStabHolz}
}}
\mbox{
\subfloat[]
{
\includegraphics[width=0.7\textwidth]{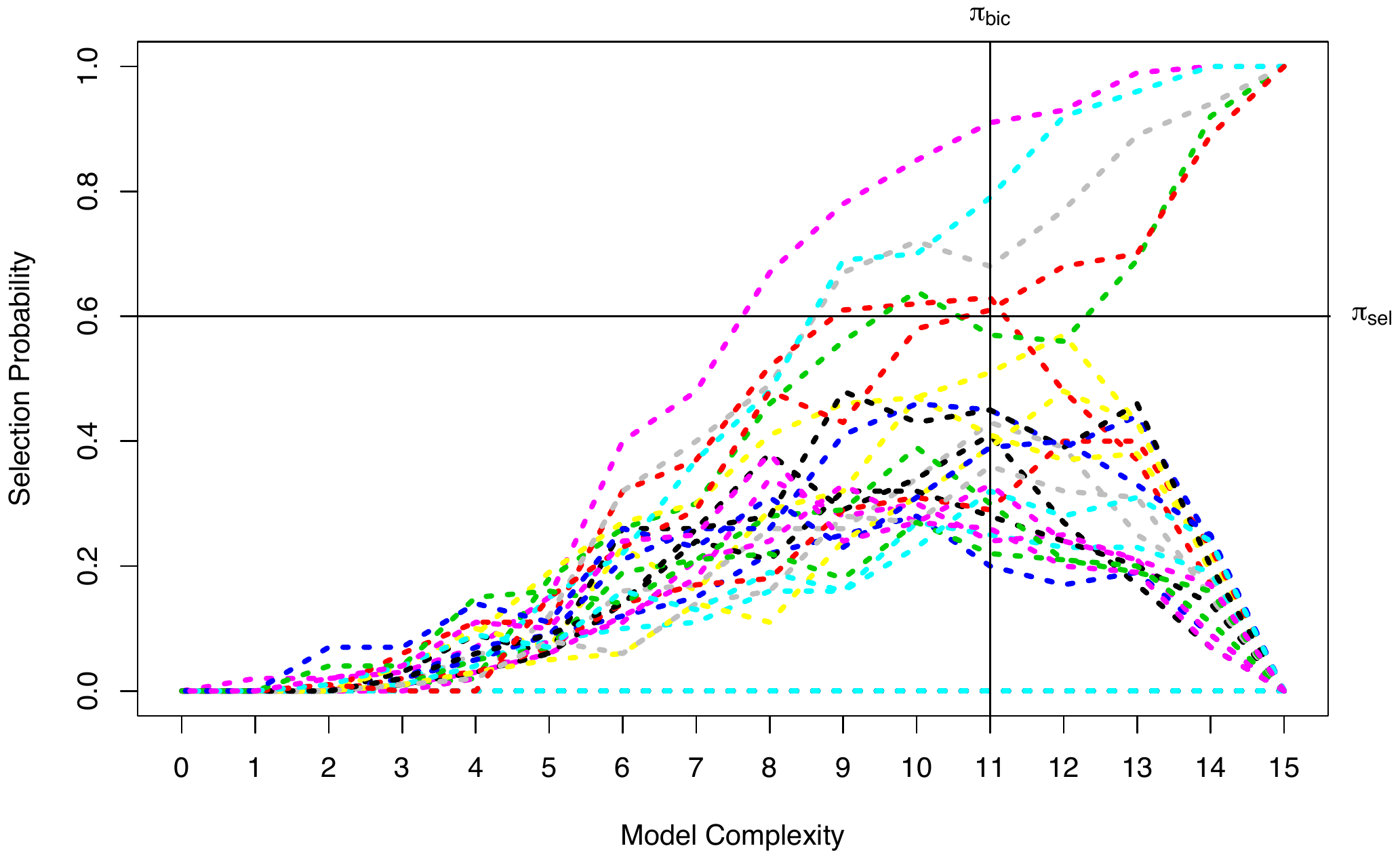}
\label{causalStabHolz} 
}}
\caption{The (a) edge and the (b) causal path stability graphs from Holzinger application. Each line indicates the frequency of relations between a pair of variables across different model complexities. The frequency is computed relative to the number of SEMs in each model complexity. The causal path only stability takes into account causal relations with any length, while the edge stability takes into account all relations regardless of the direction.}
\label{Figure:stabHolz}
\end{figure}

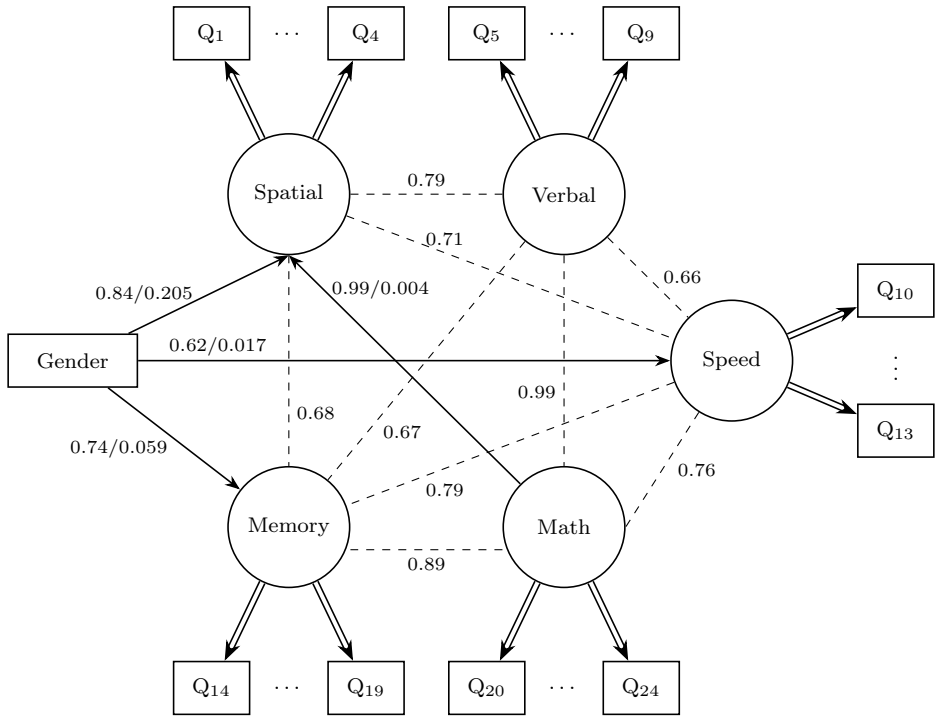
\begin{figure}[!htb]
\center
\footnotesize
\begin{tikzpicture}[
  transform shape, node distance=2cm,
  roundnode/.style={circle, draw=black, semithick, minimum size=16mm},
  squarednode/.style={rectangle, draw=black, semithick, minimum width=10mm, minimum height=7mm},
  squarednode2/.style={rectangle, draw=black, semithick, minimum width=17mm, minimum height=7mm},
  ellipsenode/.style={ellipse, draw=black, semithick, minimum size=5mm},
  arrow/.style = {semithick,-Stealth},
  dotnode/.style={fill,inner sep=0pt,minimum size=2pt,circle} 
]

\node[roundnode] (spatial) {Spatial};
\node[above=1.2cm of spatial] (qSpa) {\dots}; 
\node[squarednode, left=0.2cm of qSpa]    (q1) {Q$_1$};
\node[squarednode, right=0.2cm of qSpa]    (q4) {Q$_4$};
\draw[arrow, double distance=1.5pt] (spatial) -- node{} (q1);
\draw[arrow, double distance=1.5pt] (spatial) -- node{} (q4);

\node[roundnode, right=2cm of spatial] (verbal) {Verbal};
\node[above=1.2cm of verbal] (qVer) {\dots}; 
\node[squarednode, left=0.2cm of qVer]    (q5) {Q$_5$};
\node[squarednode, right=0.2cm of qVer]    (q9) {Q$_9$};
\draw[arrow, double distance=1.5pt] (verbal) -- node{} (q5);
\draw[arrow, double distance=1.5pt] (verbal) -- node{} (q9);

\node[roundnode, below right= 1.5cm of verbal] (speed) {Speed};
 \node[right=1.2cm of speed] (qSpe) {\vdots}; 
 \node[squarednode, below=0.2cm of qSpe]    (q13) {Q$_{13}$};
\node[squarednode, above=0.2cm of qSpe]    (q10) {Q$_{10}$};
\draw[arrow, double distance=1.5pt] (speed) -- node{} (q10);
\draw[arrow,double distance=1.5pt] (speed) -- node{} (q13);

\node[roundnode, below left= 1.5cm of speed] (math) {Math};
\node[below=1.2cm of math] (qMath) {\dots}; 
\node[squarednode, right=0.2cm of qMath]    (q24) {Q$_{24}$};
\node[squarednode, left=0.2cm of qMath]    (q20) {Q$_{20}$};
\draw[arrow, double distance=1.5pt] (math) -- node{} (q24);
\draw[arrow, double distance=1.5pt] (math) -- node{} (q20);

\node[roundnode, left= 2cm of math] (memory) {Memory};
\node[below=1.2cm of memory] (qMem) {\dots}; 
\node[squarednode, right=0.2cm of qMem]    (q19) {Q$_{19}$};
\node[squarednode, left=0.2cm of qMem]    (q14) {Q$_{14}$};
\draw[arrow, double distance=1.5pt] (memory) -- node{} (q14);
\draw[arrow, double distance=1.5pt] (memory) -- node{} (q19);

\node[squarednode2, left= 7cm of speed] (gender) {Gender};

\draw[arrow] (gender) -- node[left=0.1,font=\scriptsize]{0.84/0.205} (spatial.south);
\draw[arrow] (gender.east) -- node[above=-0.05, pos=0.15,font=\scriptsize]{0.62/0.017} (speed);
\draw[arrow] (gender) -- node[left=0.15, pos=0.6,font=\scriptsize]{0.74/0.059} (memory);
\draw[arrow] (math) -- node[right, pos=0.85,font=\scriptsize]{0.99/0.004} (spatial.south);

\draw (spatial) edge[dashed] node[above,font=\scriptsize]{0.79}(verbal);
\draw (spatial) edge[dashed] node[above,pos=0.3,font=\scriptsize]{0.71}(speed);
\draw (spatial) edge[dashed] node[pos=0.75,right,font=\scriptsize]{0.68}(memory);

\draw (verbal) edge[dashed] node[right=0.1,font=\scriptsize]{0.66}(speed);
\draw (verbal) edge[dashed] node[pos=0.8,right=0.1,font=\scriptsize]{0.67}(memory);
\draw (verbal.south) edge[dashed] node[pos=0.65, left,font=\scriptsize]{0.99}(math.north);

\draw (speed) edge[dashed] node[below=0.1,pos=0.7,font=\scriptsize]{0.79}(memory);
\draw (speed) edge[dashed] node[pos=0.5,right=0.1,font=\scriptsize]{0.76}(math.east);

\draw ([yshift=-3mm]math.west) edge[dashed] node[below,font=\scriptsize]{0.89}([yshift=-3mm]memory.east);

\end{tikzpicture}
\caption{A dashed undirected edge exhibits strong association where the causal direction cannot be determined from the data. A directed edge (arrow) with a single line indicates a causal relation and that with a double line indicates a factor loading. All edges, except the factor loadings, are annotated with a reliability score, i.e., the highest selection probability an edge has across the relevant region of the edge stability graph. In addition, all arrows are annotated with an estimate of the total causal effect (see~\ref{Section:CausalEffect} for more detail). For example, $0.62\slash 0.017$ indicates a reliability score of $0.62$ and a total causal effect of $0.017$.}
\label{Figure:modelHolz}
\end{figure}

Figure~\ref{Figure:stabHolz} displays the edge and the causal path stability graphs.
We set $\pi_{sel}=0.6$ and found that $\pi_{bic}=11$ (cf. Section~\ref{Section:discussionADHD}). Figure~\ref{Figure:modelHolz} visualizes the relevant structures into a graph.

Based on Figure~\ref{Figure:modelHolz}, we see that gender influences spatial, mental speed, and memory ability. The causal relation from gender to spatial ability matches the meta-analysis of \citep{linn1985emergence} that
indicated that males tend to perform better than females in specific types of spatial ability. The causal relation from gender to memory is in agreement with the results of \citep{cattaneo2006gender} and \citep{longman2007wais}
that exhibited gender differences in favor of males on object and word location memory, and working memory index, respectively. 
Gender was also found to influence the mental speed, as indicated by \citep{der2006age}
that pointed out a male advantage in reaction times using various measures.

It was also found that mathematical ability affects spatial ability. Previous studies indicated the association between the two \citep{burnett1979spatial,casey2001spatial,robinson1996structure}. Other studies emphasized that mathematics is one of the concepts that is mentally represented in a spatial format \citep{barsalou2008grounded,lakoff2000mathematics}. Moreover, the study by \citep{mix2016separate}
detailed specific mathematical tasks that predicts the most variance in the spatial ability of children aged $5$ to $13$.

All latent variables seem to be associated directly or indirectly. This might stem from some items on different latent variables that are possibly overlapping, for example, arithmetic (mathematical ability) and addition tasks (mental speed). Some associations are likely due to a common direct relation that a pair of variables has, for example, spatial ability and mental speed are affected by gender and both are associated.
That being said, previous studies \citep{lynn1987intelligence,cantor1991short,de2009working,daneman1980individual,vukovic2013relationship,mitolo2015relationship} indicated and discussed most of the associations.

\section{Conclusion and future work}
\label{Section:Conclusion}
It is often of great interest in many fields such as sociology, psychology, and medicine to model causal relationship among latent constructs that are indicated through observed proxies. 
In the present study, we extended S3C to S3C-Latent to model causal relations among latent variables. The chief aim of S3C-Latent is to resolve the problems of the inherent instability in model estimation and the immense number of possible models. To realize that, S3C-Latent recasts the concept of stability selection into a multi-objective optimization problem, and jointly optimizes across the whole range of model complexities, resulting in Pareto optimal models. One also can see S3C-Latent as a general framework that can be integrated with other causal discovery methods for latent variables without altering their original assumptions. In principle, it can be realized by running a causal discovery method on different data subsets and then measuring the same edge and causal path stability based on the inferred models.

We compared the results of S3C-Latent to those of PC-MIMBuild on simulated data generated with different schemes, varying the number of latent variables, sample size, number of indicators, and types of indicators (continuous and ordinal). The comparison showed that S3C-Latent performs better than PC-MIMBuild. 
Moreover, we applied S3C-Latent to real-world ADHD and Holzinger data sets. In general, the results are consistent with those of earlier studies.

The current version of S3C-Latent, however, is designed for cross-sectional data, assumes no reciprocal causal relation, and cannot handle missing values other than through imputation prior to application. Thus, future work should consider extensions to longitudinal data, reciprocal causal relations, and a more sophisticated way of handling missing values.

\section*{References}

\bibliography{mybibfile}

\begin{thebibliography}{10}
\expandafter\ifx\csname url\endcsname\relax
  \def\url#1{\texttt{#1}}\fi
\expandafter\ifx\csname urlprefix\endcsname\relax\def\urlprefix{URL }\fi
\expandafter\ifx\csname href\endcsname\relax
  \def\href#1#2{#2} \def\path#1{#1}\fi

\bibitem{silva2006learning}
R.~Silva, R.~Scheine, C.~Glymour, P.~Spirtes, Learning the structure of linear
  latent variable models, Journal of Machine Learning Research 7~(Feb) (2006)
  191--246.

\bibitem{jacobsen1999relation}
P.~Jacobsen, L.~Azzarello, D.~Hann, Relation of catastrophizing to fatigue
  severity in women with breast cancer, Cancer Res Ther Control 8~(155) (1999)
  1999--164.

\bibitem{sullivan1990relation}
M.~J. Sullivan, J.~L. D'Eon, Relation between catastrophizing and depression in
  chronic pain patients., Journal of abnormal psychology 99~(3) (1990) 260.

\bibitem{bollen1989}
K.~Bollen, Structural Equations With Latent Variables, WILEY, 1989.

\bibitem{ruifei2018copulafactor}
R.~Cui, P.~Groot, M.~Schauer, T.~Heskes, Learning the causal structure of
  copula models with latent variables, in: Conference on Uncertainty in
  Artificial Intelligence (Accepted), 2018.

\bibitem{harwood2002genetic}
S.~Harwood, R.~Scheines, Genetic algorithm search over causal models.

\bibitem{long1983covariance}
J.~S. Long, Covariance structure models: An introduction to LISREL, no.~34,
  Sage, 1983.

\bibitem{maccallum1986specification}
R.~MacCallum, Specification searches in covariance structure modeling,
  Psychological Bulletin 100~(1) (1986) 107--120.

\bibitem{rahmadi2017causality}
R.~Rahmadi, P.~Groot, M.~Heins, H.~Knoop, T.~Heskes, the
  {OPTIMISTIC}~consortium, Causality on cross-sectional data: Stable
  specification search in constrained structural equation modeling, Applied
  Soft Computing 52 (2017) 687--698.

\bibitem{spirtes2000causation}
P.~Spirtes, C.~N. Glymour, R.~Scheines, Causation, prediction, and search,
  Vol.~81, MIT press, 2000.

\bibitem{joreskog1977structural}
K.~G. Joreskog, Structural equation models in the social sciences:
  Specification estimation and testing, Applications of statistics (1977)
  265--287.

\bibitem{spirtes2010introduction}
P.~Spirtes, Introduction to causal inference, The Journal of Machine Learning
  Research 11 (2010) 1643--1662.

\bibitem{olsson1982polyserial}
U.~Olsson, F.~Drasgow, N.~J. Dorans, The polyserial correlation coefficient,
  Psychometrika 47~(3) (1982) 337--347.

\bibitem{drasgow1988polychoric}
F.~Drasgow, Polychoric and polyserial correlations, Encyclopedia of statistical
  sciences.

\bibitem{chickering2002learning}
D.~M. Chickering, Learning equivalence classes of {B}ayesian-network
  structures, The Journal of Machine Learning Research 2 (2002) 445--498.

\bibitem{kenny1998data}
D.~Kenny, D.~Kashy, N.~Bolger, Data analysis in social psychology (in d.
  gilbert, s. fiske, \& g. lindzey (eds.). the handbook of social psychology
  (vol. 1, pp. 233--265) (1998).

\bibitem{weston2006brief}
R.~Weston, P.~A. Gore~Jr, A brief guide to structural equation modeling, The
  counseling psychologist 34~(5) (2006) 719--751.

\bibitem{joreskog1978structural}
K.~G. J{\"o}reskog, Structural analysis of covariance and correlation matrices,
  Psychometrika 43~(4) (1978) 443--477.

\bibitem{deb2002fast}
K.~Deb, A.~Pratap, S.~Agarwal, T.~Meyarivan, A fast and elitist multiobjective
  genetic algorithm: {NSGA-II}, IEEE Transactions on Evolutionary Computation
  6~(2) (2002) 182--197.

\bibitem{meinshausen2010stability}
N.~Meinshausen, P.~B{\"u}hlmann, Stability selection, Journal of the Royal
  Statistical Society: Series B (Statistical Methodology) 72~(4) (2010)
  417--473.

\bibitem{joreskog1967some}
K.~G. J{\"o}reskog, Some contributions to maximum likelihood factor analysis,
  Psychometrika 32~(4) (1967) 443--482.

\bibitem{ramsey2010bootstrapping}
J.~Ramsey, Bootstrapping the {PC} and {CPC} algorithms to improve search
  accuracy.

\bibitem{fawcett2004roc}
T.~Fawcett, {ROC} graphs: Notes and practical considerations for researchers,
  Machine learning 31 (2004) 1--38.

\bibitem{van2012co}
D.~J. van Steijn, J.~S. Richards, A.~M. Oerlemans, S.~W. de~Ruiter, M.~A. van
  Aken, B.~Franke, J.~Buitelaar, N.~N. Rommelse, et~al., The co-occurrence of
  autism spectrum disorder and attention-deficit/hyperactivity disorder
  symptoms in parents of children with {ASD} or {ASD} with {ADHD}, Journal of
  Child Psychology and Psychiatry 53~(9) (2012) 954--963.

\bibitem{honaker2011amelia}
J.~Honaker, G.~King, M.~Blackwell, et~al., Amelia {II}: A program for missing
  data, Journal of statistical software 45~(7) (2011) 1--47.

\bibitem{gaub1997gender}
M.~Gaub, C.~L. Carlson, Gender differences in {ADHD}: a meta-analysis and
  critical review, Journal of the American Academy of Child \& Adolescent
  Psychiatry 36~(8) (1997) 1036--1045.

\bibitem{gershon2002meta}
J.~Gershon, J.~Gershon, A meta-analytic review of gender differences in {ADHD},
  Journal of attention disorders 5~(3) (2002) 143--154.

\bibitem{bauermeister2007adhd}
J.~J. Bauermeister, P.~E. Shrout, L.~Ch{\'a}vez, M.~Rubio-Stipec,
  R.~Ram{\'\i}rez, L.~Padilla, A.~Anderson, P.~Garc{\'\i}a, G.~Canino, {ADHD}
  and gender: are risks and sequela of {ADHD} the same for boys and girls?,
  Journal of Child Psychology and Psychiatry 48~(8) (2007) 831--839.

\bibitem{willcutt2000etiology}
E.~G. Willcutt, B.~F. Pennington, J.~C. DeFries, Etiology of inattention and
  hyperactivity/impulsivity in a community sample of twins with learning
  difficulties, Journal of Abnormal Child Psychology 28~(2) (2000) 149--159.

\bibitem{novik2006influence}
T.~S. N{\o}vik, A.~Hervas, S.~J. Ralston, S.~Dalsgaard, R.~R. Pereira, M.~J.
  Lorenzo, A.~S. Group, et~al., Influence of gender on
  attention-deficit/hyperactivity disorder in europe--adore, European Child \&
  Adolescent Psychiatry 15~(1) (2006) i15--i24.

\bibitem{andreou2005verbal}
G.~Andreou, P.~Agapitou, A.~Karapetsas, Verbal skills in children with {ADHD},
  European Journal of Special Needs Education 20~(2) (2005) 231--238.

\bibitem{frazier2004meta}
T.~W. Frazier, H.~A. Demaree, E.~A. Youngstrom, Meta-analysis of intellectual
  and neuropsychological test performance in attention-deficit/hyperactivity
  disorder., Neuropsychology 18~(3) (2004) 543.

\bibitem{sokolova2017causal}
E.~Sokolova, A.~M. Oerlemans, N.~N. Rommelse, P.~Groot, C.~A. Hartman, J.~C.
  Glennon, T.~Claassen, T.~Heskes, J.~K. Buitelaar, A causal and mediation
  analysis of the comorbidity between attention deficit hyperactivity disorder
  ({ADHD}) and autism spectrum disorder ({ASD}), Journal of autism and
  developmental disorders 47~(6) (2017) 1595--1604.

\bibitem{sokolova2016statistical}
E.~Sokolova, P.~Groot, T.~Claassen, K.~J. van Hulzen, J.~C. Glennon, B.~Franke,
  T.~Heskes, J.~Buitelaar, Statistical evidence suggests that inattention
  drives hyperactivity/impulsivity in attention deficit-hyperactivity disorder,
  PLoS one 11~(10) (2016) e0165120.

\bibitem{holzinger1939study}
K.~J. Holzinger, F.~Swineford, A study in factor analysis: The stability of a
  {B}i-factor solution., Supplementary Educational Monographs.

\bibitem{kelley2007methods}
K.~Kelley, Methods for the behavioral, educational, and social sciences: An {R}
  package, Behavior Research Methods 39~(4) (2007) 979--984.

\bibitem{douglas1940Factor}
D.~E. Scates, A study in factor analysis: The stability of a {Bi}-{Factor}
  solution. {K}arl {J}. {H}olzinger, {F}rances {S}wineford, The Elementary
  School Journal 40~(6) (1940) 468--469.
\newblock \href {http://dx.doi.org/10.1086/457783} {\path{doi:10.1086/457783}}.

\bibitem{linn1985emergence}
M.~C. Linn, A.~C. Petersen, Emergence and characterization of sex differences
  in spatial ability: A meta-analysis, Child development (1985) 1479--1498.

\bibitem{cattaneo2006gender}
Z.~Cattaneo, A.~Postma, T.~Vecchi, Gender differences in memory for object and
  word locations, Quarterly Journal of Experimental Psychology 59~(5) (2006)
  904--919.

\bibitem{longman2007wais}
R.~S. Longman, D.~H. Saklofske, T.~S. Fung, Wais-iii percentile scores by
  education and sex for us and canadian populations, Assessment 14~(4) (2007)
  426--432.

\bibitem{der2006age}
G.~Der, I.~J. Deary, Age and sex differences in reaction time in adulthood:
  results from the united kingdom health and lifestyle survey., Psychology and
  aging 21~(1) (2006) 62.

\bibitem{burnett1979spatial}
S.~A. Burnett, D.~M. Lane, L.~M. Dratt, Spatial visualization and sex
  differences in quantitative ability, Intelligence 3~(4) (1979) 345--354.

\bibitem{casey2001spatial}
M.~B. Casey, R.~L. Nuttall, E.~Pezaris, Spatial-mechanical reasoning skills
  versus mathematics self-confidence as mediators of gender differences on
  mathematics subtests using cross-national gender-based items, Journal for
  research in mathematics education (2001) 28--57.

\bibitem{robinson1996structure}
N.~M. Robinson, R.~D. Abbott, V.~W. Berninger, J.~Busse, Structure of abilities
  in math-precocious young children: Gender similarities and differences.,
  Journal of Educational Psychology 88~(2) (1996) 341.

\bibitem{barsalou2008grounded}
L.~W. Barsalou, Grounded cognition, Annu. Rev. Psychol. 59 (2008) 617--645.

\bibitem{lakoff2000mathematics}
G.~Lakoff, R.~E. N{\'u}{\~n}ez, Where mathematics comes from: How the embodied
  mind brings mathematics into being, AMC 10 (2000) 12.

\bibitem{mix2016separate}
K.~S. Mix, S.~C. Levine, Y.-L. Cheng, C.~Young, D.~Z. Hambrick, R.~Ping,
  S.~Konstantopoulos, Separate but correlated: The latent structure of space
  and mathematics across development., Journal of Experimental Psychology:
  General 145~(9) (2016) 1206.

\bibitem{lynn1987intelligence}
R.~Lynn, The intelligence of the mongoloids: A psychometric, evolutionary and
  neurological theory, Personality and individual differences 8~(6) (1987)
  813--844.

\bibitem{cantor1991short}
J.~Cantor, R.~W. Engle, G.~Hamilton, Short-term memory, working memory, and
  verbal abilities: How do they relate?, Intelligence 15~(2) (1991) 229--246.

\bibitem{de2009working}
B.~De~Smedt, R.~Janssen, K.~Bouwens, L.~Verschaffel, B.~Boets,
  P.~Ghesqui{\`e}re, Working memory and individual differences in mathematics
  achievement: A longitudinal study from first grade to second grade, Journal
  of experimental child psychology 103~(2) (2009) 186--201.

\bibitem{daneman1980individual}
M.~Daneman, P.~A. Carpenter, Individual differences in working memory and
  reading, Journal of verbal learning and verbal behavior 19~(4) (1980)
  450--466.

\bibitem{vukovic2013relationship}
R.~K. Vukovic, N.~K. Lesaux, The relationship between linguistic skills and
  arithmetic knowledge, Learning and Individual Differences 23 (2013) 87--91.

\bibitem{mitolo2015relationship}
M.~Mitolo, S.~Gardini, P.~Caffarra, L.~Ronconi, A.~Venneri, F.~Pazzaglia,
  Relationship between spatial ability, visuospatial working memory and
  self-assessed spatial orientation ability: a study in older adults, Cognitive
  processing 16~(2) (2015) 165--176.

\bibitem{maathuis2009estimating}
M.~H. Maathuis, M.~Kalisch, P.~B{\"u}hlmann, et~al., Estimating
  high-dimensional intervention effects from observational data, The Annals of
  Statistics 37~(6A) (2009) 3133--3164.

\bibitem{pearl2000causality}
J.~Pearl, Causality: models, reasoning and inference, Vol.~29, Cambridge
  University Press, 2000.

\bibitem{ghahramani1996algorithm}
Z.~Ghahramani, G.~E. Hinton, et~al., The {EM} algorithm for mixtures of factor
  analyzers, Tech. rep., Technical Report CRG-TR-96-1, University of Toronto
  (1996).

\end{thebibliography}
\newpage

\appendix
\section{Stability graphs}
\label{Section:StabilityGraph}
\begin{definition}
(Stability graphs \citep{rahmadi2017causality}) Let $A$ and $B$ be two variables and $G$ a multiset (or bag) of CPDAGS.  Let $G_c$ be the submultiset of $G$ containing all CPDAGS with complexity $c$. The edge stability for $A$ and $B$ at complexity $c$ is the number of models in $G_c$ for which there exists an edge between $A$ and $B$ (i.e., $A\rightarrow B$, $B\rightarrow A$, or $A-B$) divided by the total number of models in $G_c$. The causal path stability for $A$ to $B$ at complexity $c$ is the number of models in $G_c$ for which there is a directed path from $A$ to $B$ (of any length) divided by the total number of models in $G_c$. The terms edge stability graph and causal path stability graph are used to denote the corresponding measures for all variable pairs and all complexity levels.
\end{definition}

\section{Comparison of S3C-Latent results}
\label{Section:SchemeComp}
\begin{figure}[!h]
\centering
\includegraphics[width=1\textwidth]{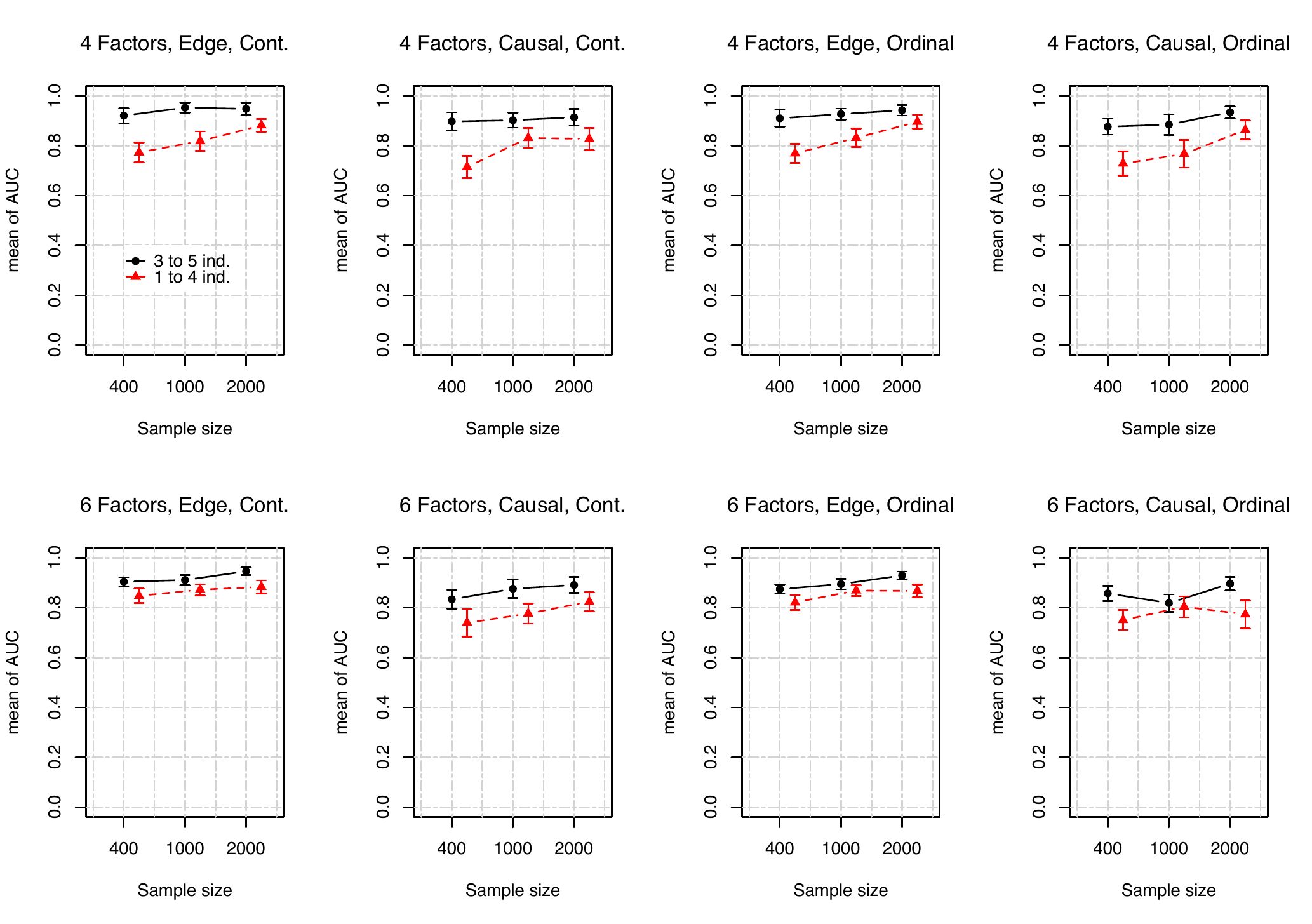} 
\caption{Comparison between the mean AUC obtained by S3C-Latent on the data generated from SEMs with $3$ to $5$ indicators per latent variable to those obtained on data generated from SEMs with $1$ to $4$ indicators per latent variable.
The plots in the top panel are results of simulation on the data generated from SEMs of $4$ latent variables and those in the bottom panel are results of simulation on the data sets generated from SEMs of $6$ latent variables.}
\label{Figure:SchemeABS3C}
\end{figure}

\section{Causal effect estimation}
\label{Section:CausalEffect}
The procedure originates from IDA method\citep{maathuis2009estimating}. 
Let $G=\{G_1,\dotsc,G_m\}$ be a CPDAG to which $m$ DAGs belong to. For each DAG $G_j$, IDA administers intervention calculus \cite{pearl2000causality} to obtain multisets $\Theta_i=\{\theta_{ij}\}_{j \in{1,\dotsc,m}}, i=1,\dotsc,p$, where $p$ is the number of covariates. In particular,  $\theta_{ij}$ defines the possible causal effect of $X_i$ on $Y$ in graph $G_j$. Assuming that the data have been drawn from a linear Gaussian model, the causal effects of  $X_i$ on $Y$ can be estimated through a regression of $Y$ on $X_i$ and its parents. 
Note that the possible causal effect $\theta_{ij}$ can be direct, indirect, or addition of the both effects, depending on the structure of $G_j$.

As S3C and S3C-Latent result in many CPDAGs representing optimal models, we extend IDA as follows.  We particularly use $G_{\pi_{\mathrm{bic}}}$, the CPDAGs of all optimal models with complexity equal to $\pi_{\mathrm{bic}}$. For each CPDAG $G\in G_{\pi_{\mathrm{bic}}}$, we compute the possible causal effects of each relevant causal path, e.g., $X$ on $Y$, obtaining estimates $\Theta_{X\to Y}^k, k=1,\dotsc,N$, where $N$ is the number of data subsets. We then concatenate all causal effect estimates in $\Theta_{X\to Y}^k$ into a single multiset $\Theta_{X\to Y}$.

Finally, we use the median $\tilde{\Theta}_{X\to Y}$ to represent the estimated total causal effects from $X$ to $Y$. In the case of both $X$ and $Y$ are continuous, $\tilde{\Theta}_{X\to Y}$ is standardized via
$\tilde{\Theta}_{X\to Y}\times \sigma_X/\sigma_Y,$
where $\sigma_X$ and $\sigma_Y$ are the standard deviations of the covariate and the response, respectively.

In order to estimate causal effect from a SEM with latent variables, we need to sample data from the latent variables. The steps to sample from latent variables are described in~\ref{Section:SamplingLatent}.

\section{Sampling from latent variable}
\label{Section:SamplingLatent}
In what follows, we describe the steps for the sampling. Suppose we are given a measurement model that reads
\begin{equation}
	\bm{x}=\bm{\Lambda}\bm{\eta}+\bm{\epsilon},
\end{equation}
where $\bm\eta$ is a vector of latent (for simplicity, can be either endogenous or exogenous) variables, $\bm\Lambda$ is a matrix of factor loadings, $\bm{x}$ is a vector of indicators, $\bm{\epsilon}$ contain errors of the indicators, and $\bm{\Theta}$ is a covariance matrix of $\bm{\epsilon}$ and is diagonal.

Following \citep{ghahramani1996algorithm},
given $\bm\Lambda$ and $\bm\Theta$, the expected value of latent variables $\bm{\eta}$ can be computed via the linear projection:
\begin{equation}
	\mathbb{E}(\bm\eta|\bm{x})=\beta\bm{x},
\end{equation}
with $\beta \equiv \bm\Lambda'(\bm\Theta+\bm{\Lambda\Lambda}')^{-1}$, and the variance of $\bm\eta$ can be computed through,
\begin{equation}
	\mathrm{Var}(\bm\eta|\bm{x})=I-\beta\bm\Lambda.
\end{equation}

We can then sample $\hat{\bm\eta}\sim \mathcal{N}(\bm\mu,\,\bm\sigma^{2})$, where $\bm\mu$ is the mean of vector $\mathbb{E}(\bm\eta|\bm{x})$ and $\bm\sigma^{2}=\mathrm{Var}(\bm\eta|\bm{x})$.

\end{document}